\documentclass[conference]{IEEEtran}
\usepackage{amsmath,amssymb,graphicx,cite,booktabs,array,tabularx,multirow}
\usepackage[table]{xcolor}
\usepackage[hidelinks]{hyperref}
\usepackage{xurl,placeins}
\usepackage{pifont}
\usepackage{iftex}
\ifPDFTeX
  \usepackage[T1]{fontenc}
  \usepackage[utf8]{inputenc}
  \usepackage{tgtermes,tgheros,tgcursor}
\else
\usepackage{fontspec}
\fi

\begin{document}
\title{CAFE: Closed-Loop Autonomous Task Formulation and Execution without Instance-Specific Instructions}
\hypersetup{pdfauthor={Mingjia Chen, Xuewen Zhao, Philip Dames, Qi Mao, Fei Xie, Jun Chen}}
\author{\IEEEauthorblockN{Mingjia Chen\textsuperscript{1}, Xuewen Zhao\textsuperscript{1},
Philip Dames\textsuperscript{2}, Qi Mao\textsuperscript{1}, Fei Xie\textsuperscript{1},
and Jun Chen\textsuperscript{1,*}}%
\IEEEauthorblockA{\footnotesize
\textsuperscript{1}School of Electrical and Automation Engineering, Nanjing Normal University, Nanjing, Jiangsu 210023, China\\
\textsuperscript{2}Department of Mechanical Engineering, Temple University, Philadelphia, PA 19122, USA\\
\textsuperscript{*}Corresponding author: \href{mailto:jun.chen@nnu.edu.cn}{jun.chen@nnu.edu.cn}}}
\maketitle
\begin{abstract}
Scene-driven robot autonomy requires selecting tasks from observations as well as executing them. We study scene-driven task formulation and execution under predefined cleaning, safety, and storage/organization categories, without instance-specific goals or action sequences. CAFE maintains online memory of objects, functional zones, and execution states. A rule-guided Trigger-LLM admits formulation from accumulated evidence; Draft, Complete, and Audit propose goals, resolve execution details, and check dependencies. Outcomes update memory for correction and subsequent discovery. Evaluation in 12 AI2-THOR scenes compares three backbones and task-complexity tiers using coverage, constraint, and execution metrics. Tri-LLM records higher GCR scores than the evaluated baselines across three backbones and task-complexity tiers. Controls examine completion, auditing, zones, feedback, triggering, and prompt examples. LimoCobot traces illustrate storage, cleanup, and recovery from a failed grasp. The study examines how shared scene state connects task discovery to execution and outcome-driven revision. Project page: \url{https://github.com/submissionforpaper/CAFE-Closed-Loop-Autonomous-Task-Formulation-and-Execution-without-Instance-Specific-Instructions}.
\end{abstract}
\begin{figure*}[t]
\centering
\includegraphics[trim=0bp 23.7bp 0bp 12.2bp,clip,width=0.76\textwidth]{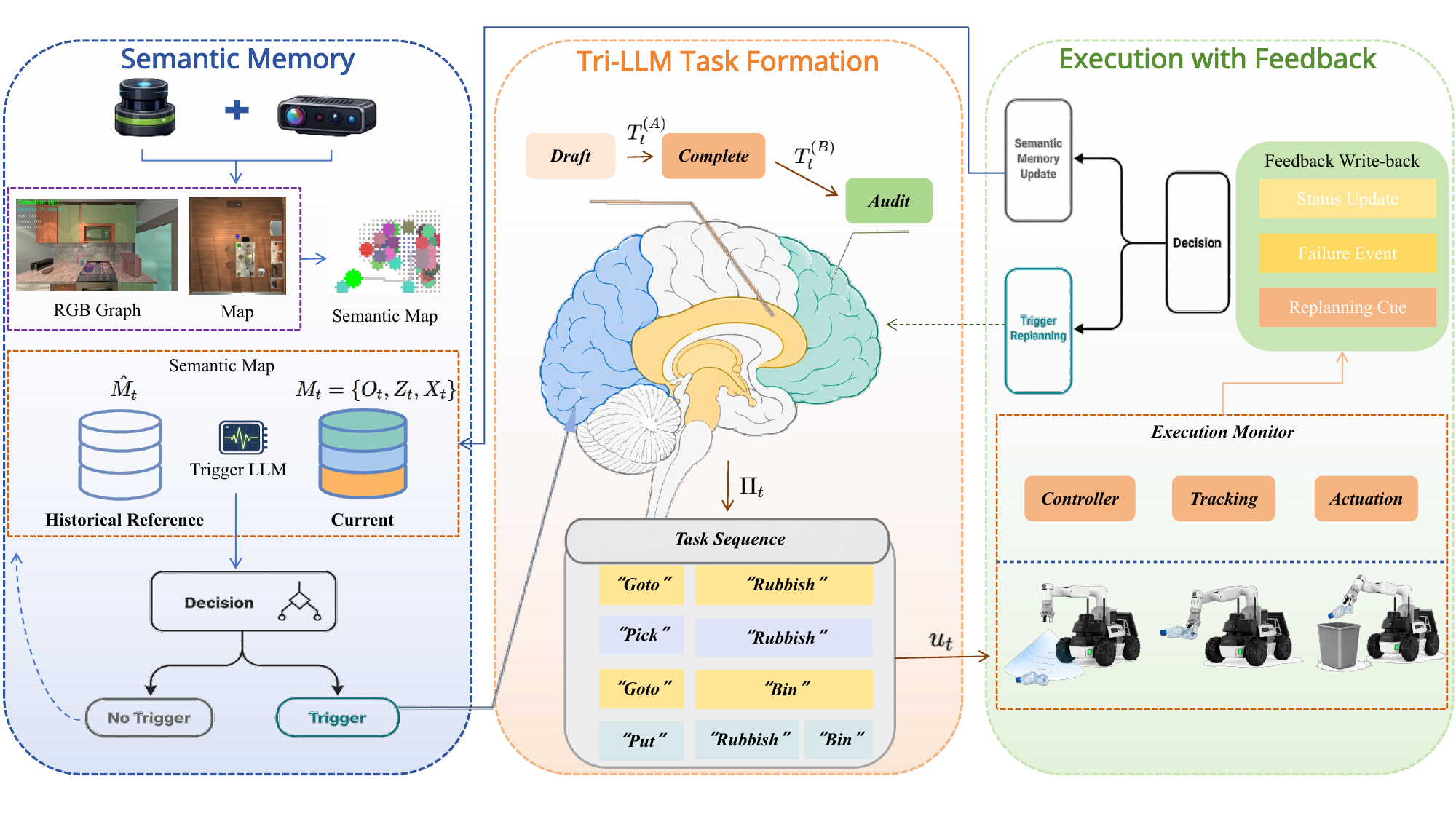}
\caption{Overview of CAFE. Online observations update semantic memory; Trigger-LLM admits a Draft--Complete--Audit reasoning pass to form ordered executable tasks; execution outcomes return to memory for correction and replanning.}
\label{fig:overall_pipeline}
\end{figure*}
\section{Introduction}
Language and vision-language models have extended robotic planning to goal-directed navigation, open-vocabulary manipulation, and long-horizon execution~\cite{chang2023goat,liu2024ok,huang2023voxposer,mon2025embodied}. PaLM-E and SayCan connect supplied goals to observations and executable skills~\cite{driess2023palm,ahn2022can}. Exploration can recover missing information for an instruction~\cite{wu2025embodied}, while task-discovery methods identify opportunities from environmental and human cues~\cite{song2025human}. A robot operating without scene-specific instructions must additionally decide which observed conditions call for action and when enough evidence is available to formulate a task.

This setting requires a recurrent connection between exploration, formulation, and execution. An observation of a dirty bowl may suggest washing, but an executable plan also needs a sink, tools, preparation steps, and dependencies. Those bindings must remain available as the robot moves, and observed action outcomes must update the state used to select subsequent tasks. Persistent scene representations provide spatial and semantic context~\cite{zhang2025mapnav,gadre2022continuous}; observation filtering helps select useful evidence~\cite{gupta2026scale}. CAFE connects that evidence to concrete proposals, resolves their physical parameters and dependencies, and revises behavior after execution.

Our objective is to connect scene-supported task discovery, executable-plan formation, and outcome-driven revision in a recurrent loop. CAFE addresses this objective under predefined cleaning, safety, and storage/organization categories. The robot receives no instance-specific goal, target-object list, or action sequence. Online semantic memory stores objects, functional zones, and execution states. Trigger-LLM admits formulation from accumulated evidence, and a staged Tri-LLM pipeline separates Draft, Complete, and Audit. Execution outcomes update the same memory for recovery and subsequent task discovery.

The contributions are:
\begin{itemize}
\item A memory-centered autonomy loop linking scene-driven task discovery to execution-state updates without instance-specific human task instructions.
\item A mechanism for recurrent task formulation from scene state, implemented through selective admission, task-status context, and Draft--Complete--Audit reasoning.
\item Evaluation across three backbones and complexity tiers, with controls for formulation, feedback, triggering, and prompt examples, plus physical storage, cleanup, and grasp-recovery demonstrations.
\end{itemize}

\section{Related Work}
\subsection{From Supplied Goals to Task Discovery}
PaLM-E and SayCan ground supplied goals in perception and skills~\cite{driess2023palm,ahn2022can}. GOAT, OK-Robot, and VoxPoser extend navigation and manipulation under known goals~\cite{chang2023goat,liu2024ok,huang2023voxposer}. ProgPrompt organizes actions and objects with programmatic preconditions~\cite{singh2023progprompt}; LLM-Planner uses grounded planning and replanning~\cite{song2023llm}, while EMPOWER integrates multiple planning roles with online grounding and execution~\cite{argenziano2024empower}. Plan-and-Act separates a Planner and Executor for goal-directed long-horizon web agents~\cite{erdogan2025plan}. These approaches organize execution around supplied goals; CAFE formulates its goals from maintained scene evidence.

AutoRT proposes scene-based activities for robot data collection~\cite{ahn2024autort}. HOTD/CMAST discovers tasks that reduce human effort across plausible futures~\cite{song2025human}; AnomalyGen addresses proactive hazard detection and resolution~\cite{song2025hazards}. These works support autonomous opportunity discovery. CAFE couples discovery to maintained scene state and completed-task context so that execution outcomes affect subsequent goal selection and de-duplication.

\subsection{Grounding and Checking Executable Plans}
Instruction following in unknown environments uses exploration to complete missing task-relevant information~\cite{wu2025embodied}. Scale-Plan filters object-rich observations before high-level reasoning~\cite{gupta2026scale}. FATE audits generated robotic curricula and repairs infeasible tasks~\cite{wei2026fate}; GraphPlan combines task and scene graphs for structured planning and event-driven replanning~\cite{li2026graphplan}. Plan verification critiques and refines erroneous action sequences~\cite{hariharan2025plan}. SPCA couples LLM-generated symbolic descriptions with PDDL planning and code validation~\cite{pesjak2026robot}. CAFE audits object availability and dependencies over memory-grounded action tuples. CAFE separates scene-supported proposal, parameter completion, and dependency auditing, admitting these stages selectively as new opportunities accumulate.

\subsection{Persistent State and Execution Feedback}
Semantic mapping organizes persistent spatial and object information for embodied reasoning~\cite{raychaudhuri2025semantic}; continuous scene models and MapNav provide concrete representations~\cite{gadre2022continuous,zhang2025mapnav}. SaPaVe jointly learns semantic camera control and active-view manipulation~\cite{liu2026sapave}. Multimodal feedback supports execution correction in changing environments~\cite{mon2025embodied}. CAFE uses object-centric memory as the shared interface for reasoning admission, task formulation, and outcome updates. Updated memory connects active-plan recovery with subsequent task discovery.
\begin{table*}[!t]
\caption{Comparison of reported system capabilities}
\label{tab:capability_comparison}
\centering
\begingroup
\definecolor{capInk}{HTML}{263B49}
\definecolor{capGroup}{HTML}{E7EDF2}
\definecolor{capHeader}{HTML}{F2F5F7}
\definecolor{capStripe}{HTML}{F5F7F9}
\definecolor{capOurs}{HTML}{E7F1F6}
\definecolor{capYes}{HTML}{24566D}
\definecolor{capPartial}{HTML}{766343}
\definecolor{capNone}{HTML}{79828A}
\newcommand{\capY}{\textcolor{capYes}{\ding{52}}}
\newcommand{\capP}{\textcolor{capPartial}{$\triangle$}}
\newcommand{\capN}{\textcolor{capNone}{$\times$}}
\newcommand{\capHead}[1]{{\sffamily\bfseries\color{capInk}#1}}
\fontsize{8.5}{10}\selectfont
\setlength{\tabcolsep}{2.5pt}
\renewcommand{\arraystretch}{1.2}
\arrayrulecolor{capInk!55}
\begin{tabularx}{\textwidth}{@{}>{\raggedright\arraybackslash}p{1.48in}*{8}{>{\centering\arraybackslash}X}@{}}
\toprule
\rowcolor{capGroup}
\capHead{Method} & \capHead{Initiation} & \multicolumn{3}{c}{\cellcolor{capGroup}\capHead{State and planning}} & \multicolumn{3}{c}{\cellcolor{capGroup}\capHead{Closed-loop operation}} & \capHead{Hardware} \\
\rowcolor{capHeader}
 & \capHead{No instance\newline instruction} & \capHead{Long-horizon\newline execution} & \capHead{Online\newline semantic\newline memory} & \capHead{Explicit\newline checks} & \capHead{Execution\newline recovery} & \capHead{Feedback\newline into new\newline goals} & \capHead{On-demand\newline reasoning} & \capHead{Real\newline robot} \\
\midrule
SayCan~\cite{ahn2022can} & \capN & \capY & \capN & \capP & \capP & \capN & \capN & \capY \\
\rowcolor{capStripe}
ProgPrompt~\cite{singh2023progprompt} & \capN & \capY & \capP & \capY & \capY & \capN & \capP & \capY \\
LLM-Planner~\cite{song2023llm} & \capN & \capY & \capY & \capN & \capY & \capN & \capY & \capN \\
\rowcolor{capStripe}
EMPOWER~\cite{argenziano2024empower} & \capN & \capY & \capP & \capN & \capN & \capN & \capN & \capY \\
AutoRT~\cite{ahn2024autort} & \capY & \capN & \capP & \capY & \capP & \capP & \capN & \capY \\
\rowcolor{capStripe}
CMAST / HOTD~\cite{song2025human} & \capY & \capN & \capN & \capY & \capN & \capN & \capN & \capN \\
AnomalyGen~\cite{song2025hazards} & \capY & \capY & \capP & \capP & \capN & \capN & \capN & \capN \\
\rowcolor{capStripe}
FATE~\cite{wei2026fate} & \capY & \capY & \capP & \capY & \capY & \capN & \capY & \capN \\
\midrule
\rowcolor{capOurs}
\textbf{\textcolor{capInk}{CAFE (ours)}} & \capY & \capY & \capY & \capY & \capY & \capY & \capY & \capY \\
\bottomrule
\end{tabularx}
\vspace{5pt}
\begin{minipage}{\textwidth}
\fontsize{8}{9.3}\selectfont\raggedright
\textit{Notes:} \capY\ reported; \capP\ partial or different scope; \capN\ unreported or out of scope. Feedback into new goals requires execution outcomes and task status to inform subsequent discovery and de-duplication through persistent memory. CAFE uses predefined discovery categories and demonstrates closed-loop execution on LimoCobot.
\end{minipage}
\endgroup
\arrayrulecolor{black}
\end{table*}

Table~\ref{tab:capability_comparison} compares reported capabilities under the stated definitions. CAFE emphasizes recurrent task formulation from scene state: frozen-reference admission selects reasoning rounds, while updated memory and task status guide subsequent goals (Sec.~\ref{sec:method}). Draft--Complete--Audit implements proposal, completion, and checking within this mechanism. The table describes original systems; adapted ProgPrompt shares CAFE's triggering, execution, and feedback infrastructure to compare formulation strategies.

\section{Method}
\label{sec:method}
Fig.~\ref{fig:overall_pipeline} guides the method: the left module maintains memory and admits reasoning; the center converts opportunities into an ordered task sequence; the right executes it and returns outcomes to memory for correction and subsequent discovery.
\subsection{Semantic State and Autonomy Scope}
\label{subsec:autonomy_scope}
\begin{figure*}[t]
\centering
\includegraphics[trim=0bp 74.8bp 0bp 112.3bp,clip,width=0.86\textwidth]{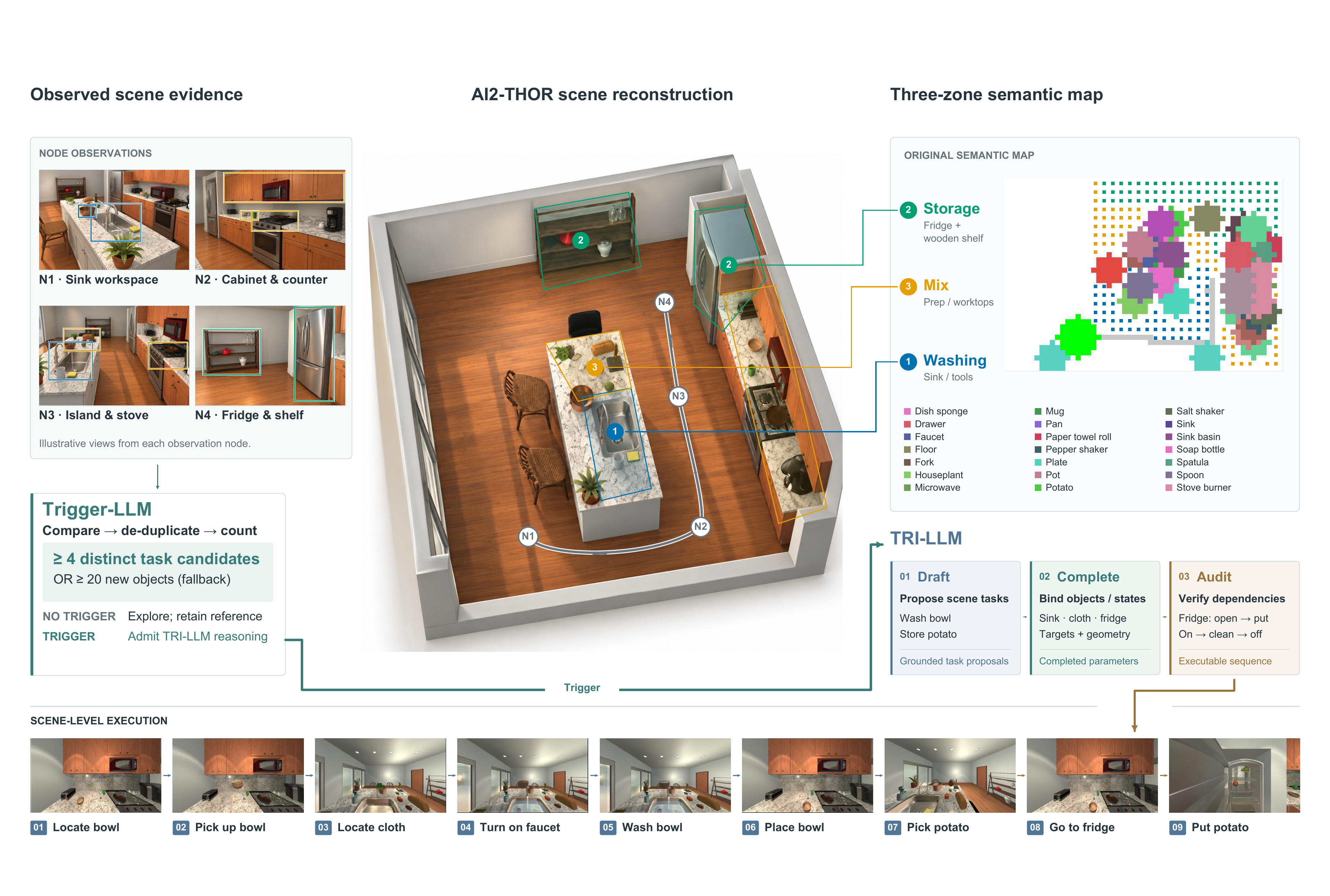}
\caption{Scene-grounded formulation. Nodes N1--N4 ground washing, storage, and preparation zones. Draft proposes bowl-washing and potato-storage tasks; Complete binds parameters, and Audit checks dependencies. The bottom strip connects the grounded tasks to selected execution actions.}
\label{fig:scene_grounding}
\end{figure*}

The memory module in Fig.~\ref{fig:overall_pipeline} operates under a fixed discovery prompt specifying cleaning, safety, and storage/organization. No scene-specific goal, target list, or action sequence is supplied. For $t\ge1$, current observations and preceding action outcomes form evidence $D_t$:
\begin{equation}
M_t=\Phi(M_{t-1},D_t),\quad
M_t=\{\mathcal O_t,\mathcal Z_t,\mathcal X_t\},\quad M_0=\emptyset.
\label{eq:memory}
\end{equation}
$\mathcal O_t$ stores grounded objects and states, $\mathcal Z_t$ functional zones, and $\mathcal X_t$ timestamps, confidence, and execution status. The maintained $M_t$ supplies a shared state interface for triggering, formulation, and feedback. A zone's landmark bounding volumes are projected onto the ground plane; their minimum and maximum coordinates define $B_z=(x_{\min},x_{\max},y_{\min},y_{\max})$. Objects inside this footprint are assigned to the zone; otherwise assignment remains deferred until further spatial evidence is available. In Fig.~\ref{fig:scene_grounding}, sink/tools support washing, refrigerator/shelf storage, and worktops preparation. Zones group objects by their functional workspaces.

\subsection{Selective Discovery Admission}
At each heartbeat, the Trigger-LLM in Fig.~\ref{fig:overall_pipeline} compares $M_t$ with a frozen reference $\hat M_t$, initialized as $\hat M_1=M_0$. The comparison includes new objects, object--zone associations, state or affordance changes, and execution outcomes. Context $C_t$ records ongoing/completed tasks and outcomes. Prompt $P_{\mathrm{trig}}$ instructs the model to identify scene-supported candidates, remove equivalent cues, and return a decision, distinct-candidate count, and brief reason:
\begin{equation}
(\tilde r_t,\hat K_t)=\Psi_{\mathrm{LLM}}(M_t,\hat M_t,C_t,P_{\mathrm{trig}}).
\label{eq:trigger}
\end{equation}
$\tilde r_t$ is the recommendation and $\hat K_t$ the estimated distinct-candidate count. The actual decision $r_t$ is Trigger if $\hat K_t\ge4$ or at least 20 objects are new relative to $\hat M_t$, and No Trigger otherwise:
\begin{equation}
\hat M_{t+1}=\begin{cases}M_t,&r_t=\mathrm{Trigger},\\\hat M_t,&\text{otherwise}.\end{cases}
\label{eq:reference}
\end{equation}
Between formulation rounds, execution/exploration continues and the reference stays frozen, accumulating evidence across observations. The trigger admits high-level task candidates whose execution parameters are resolved by the subsequent formulation stages. Failure correction bypasses this discovery gate. The controls compare triggering policies and heartbeat intervals using the admission rules above (Sec.~\ref{subsec:trigger_eval}).

\subsection{Staged Task Formulation}
The central module in Fig.~\ref{fig:overall_pipeline} grounds admitted candidates through three reasoning roles:
\begin{align}
\mathcal T_t^{(A)}&=\operatorname{DraftLLM}(M_t,C_t),\\
\mathcal T_t^{(B)}&=\operatorname{CompleteLLM}(\mathcal T_t^{(A)},M_t),\\
\Pi_t&=\operatorname{AuditLLM}(\mathcal T_t^{(B)},M_t).
\label{eq:trillm}
\end{align}
The three operators implement task proposal, completion, and auditing. Draft proposes $\tau_j=\langle a_j,\mathrm{target}_j,z_j,\mathrm{pred}_j,p_j,g_j\rangle$: action, target, zone, prerequisites, priority, and geometry. Draft is instructed to reference entities and zones in $M_t$ and flag unresolved geometric bindings.

Complete receives the flagged draft in a single forward pass. Its prompt requires resolving targets, spatial associations, and geometry from memory. The retained structure is the high-level goal set and inter-goal dependencies; Complete may add atomic navigation, preparation, and manipulation steps to realize these goals, without proposing new goals or unobserved entities. Audit is prompted to check semantic consistency, object availability, safety, and predecessor coherence, then repair recoverable inconsistencies or reject unsupported candidates. Ordering prioritizes safety, task completeness, and human-aware preferences subject to prerequisites. These roles preserve the goal layer while refining its execution layer. For potato storage, Complete expands the goal into navigation, pickup, refrigerator opening, and insertion; Audit checks the opening-before-insertion dependency. Task status in $C_t$ carries ongoing and completed goals into subsequent trigger and Draft calls, connecting action outcomes to goal de-duplication.

Fig.~\ref{fig:scene_grounding} grounds washing, storage, and preparation zones from N1--N4. The bowl is bound to a sink and cloth; washing requires faucet activation followed by shutoff. The action frames connect these bindings to execution.

\subsection{Execution and Feedback}
\label{subsec:execution_feedback}
The right module in Fig.~\ref{fig:overall_pipeline} instantiates $\Pi_t$ and monitors targets, prerequisites, and expected effects:
\begin{equation}
u_t=\pi_{\mathrm{exec}}(\Pi_t,M_t),\quad e_t^{\mathrm{fb}}=\operatorname{Monitor}(y_t^+,\Pi_t,M_t).
\label{eq:feedback}
\end{equation}
$u_t$ denotes executor output, $y_t^+$ post-action observations, and $e_t^{\mathrm{fb}}$ status/state mismatch. The executor dispatches structured action, target, zone, prerequisite, and geometric fields in the audited order. Successful outcomes are recorded in $\mathcal X_t$ before advancing to the next item. An unavailable target, violated prerequisite, or unexpected effect halts the affected sequence and requests correction. Both routine outcomes and failure events update $D_{t+1}$ and $M_{t+1}$ before the next decision. Subsequent discovery also uses completed-task context to avoid repeating fulfilled goals. This closes the feedback path in Fig.~\ref{fig:overall_pipeline}: local recovery and later discovery share updated evidence.

\section{Experiments}
\label{sec:experiments}
\subsection{Setup and Evaluation}
We evaluate 12 AI2-THOR scenes~\cite{kolve2017ai2thor} spanning bathrooms, living rooms, and kitchens, with ten repetitions per task. Simple tasks have short, largely independent chains, such as faucet shutoff. Compound tasks coordinate related goals, such as storing scattered items. Complex tasks combine longer horizons, dense dependencies, and potential failures, such as cabinet closure, washing, and kitchen organization. Semantic memory starts empty and is built through exploration. Category priors and skills are fixed, while scene-specific target lists, reference action sequences, and complete layouts are withheld. No method is trained or fine-tuned for this evaluation.

\FloatBarrier
\begin{figure*}[!t]
\centering
\includegraphics[trim=0bp 2.0bp 0bp 177.5bp,clip,width=0.94\textwidth]{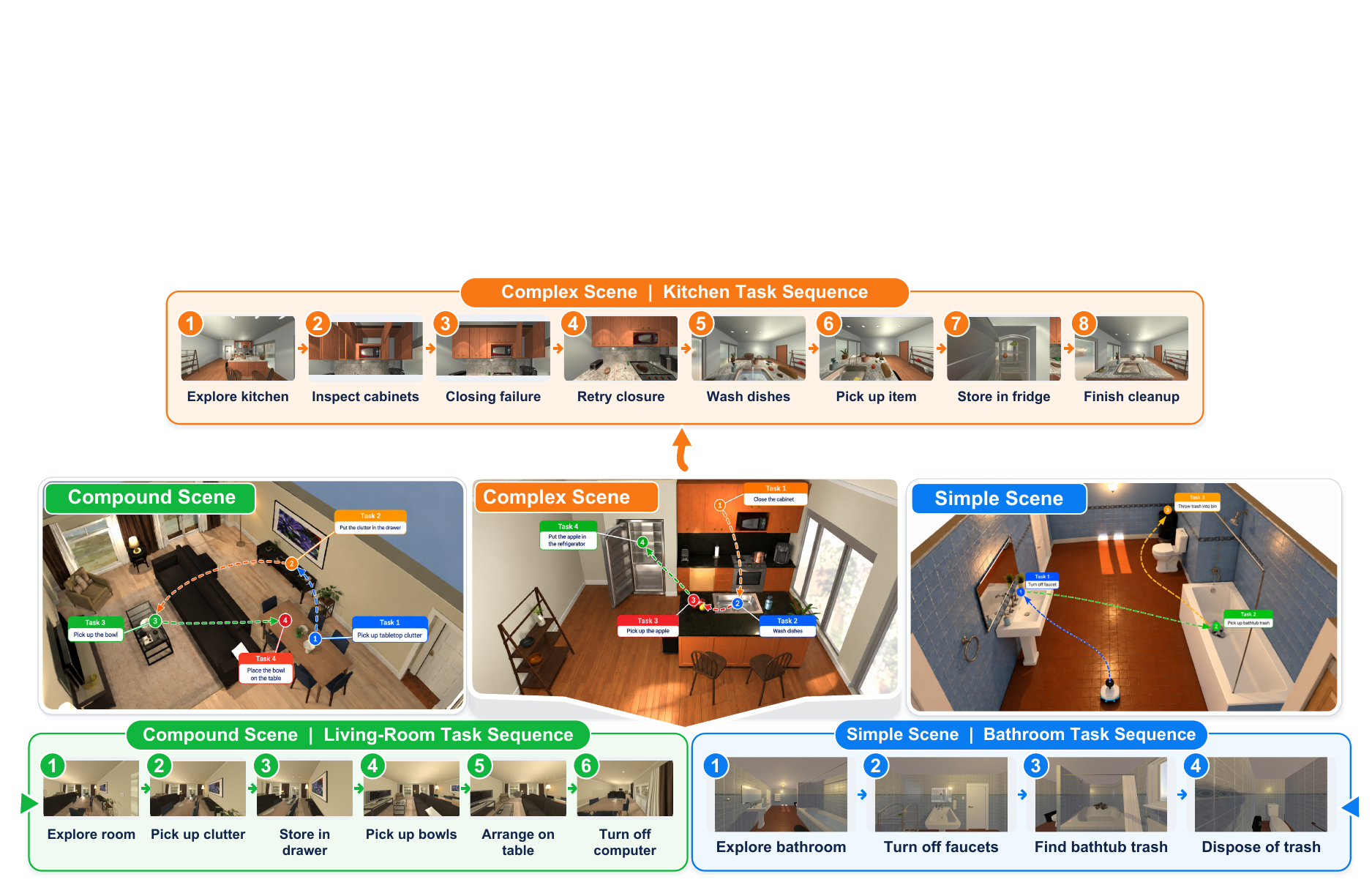}
\caption{AI2-THOR trajectories: cabinet-closing failure and retry before washing/storage (kitchen); clutter storage, bowl relocation, and computer shutoff (living room); faucet shutoff and trash disposal (bathroom). Frames show pivotal actions.}
\label{fig:simulation}
\end{figure*}

Tri-LLM and Single-LLM use DeepSeek-V3, Qwen3-Max, and Gemini-2.5-Pro. Controls use Qwen3-Max. Single-LLM uses one planner for sequence generation instead of separate Draft and Complete roles. The w/o Audit control removes Audit from the corresponding architecture. Adapted ProgPrompt~\cite{singh2023progprompt} receives semantic memory in place of its explicit task instruction. All variants share the observation interface, action space, output format, executor, triggering and feedback infrastructure, step budget, and evaluation protocol. These shared settings align the comparison of task-formulation strategies.

For instance $i$, let $P_i^\star,P_i$ be reference/inferred high-level subtask sets, $A_i^\star,A_i$ reference/inferred atomic-step sets, and $H_i$ reference hard constraints (precedence, dependencies, and exclusions). Reference sets are withheld from the robot. With nonempty denominators,
\begin{align}
\mathrm{UC}_i&=\frac{|P_i\cap P_i^\star|}{|P_i^\star|}, &
\mathrm{sUC}_i&=\frac{|A_i\cap A_i^\star|}{|A_i^\star|},\\
\mathrm{CUR}_i&=\frac{\sum_{h\in H_i}\mathbf1[h\text{ satisfied}]}{|H_i|}, &
\mathrm{SR}_i&=\frac{N_{\mathrm{planned},i}^{\mathrm{succ}}}{N_{\mathrm{planned},i}}.
\label{eq:metrics}
\end{align}
SR divides successfully executed actions by planned actions. Reported ratios are means; GCR is $\overline{\mathrm{UC}}\,\overline{\mathrm{SR}}$, rounded to three decimals. UC and sUC measure high-level goal and atomic-step coverage, respectively; CUR assesses constraint satisfaction. Together with SR and GCR, these metrics characterize task discovery, plan organization, action completeness, and grounded execution (Tables~\ref{tab:main_results} and~\ref{tab:controlled_results}). For example, discovering three of four reference subtasks gives UC $=3/4$; satisfying two of four constraints gives CUR $=2/4$; specifying ten of 21 required atomic steps gives sUC $=10/21$.

\subsection{Task Formulation and Execution Performance}
\label{subsec:main_results}
Fig.~\ref{fig:simulation} illustrates scene-derived discovery and coordination in bathroom, living-room, and kitchen trajectories across the three complexity tiers.

\begin{figure}[!t]
\centering
\includegraphics[trim=0bp 0.0bp 0bp 6.1bp,clip,width=\columnwidth]{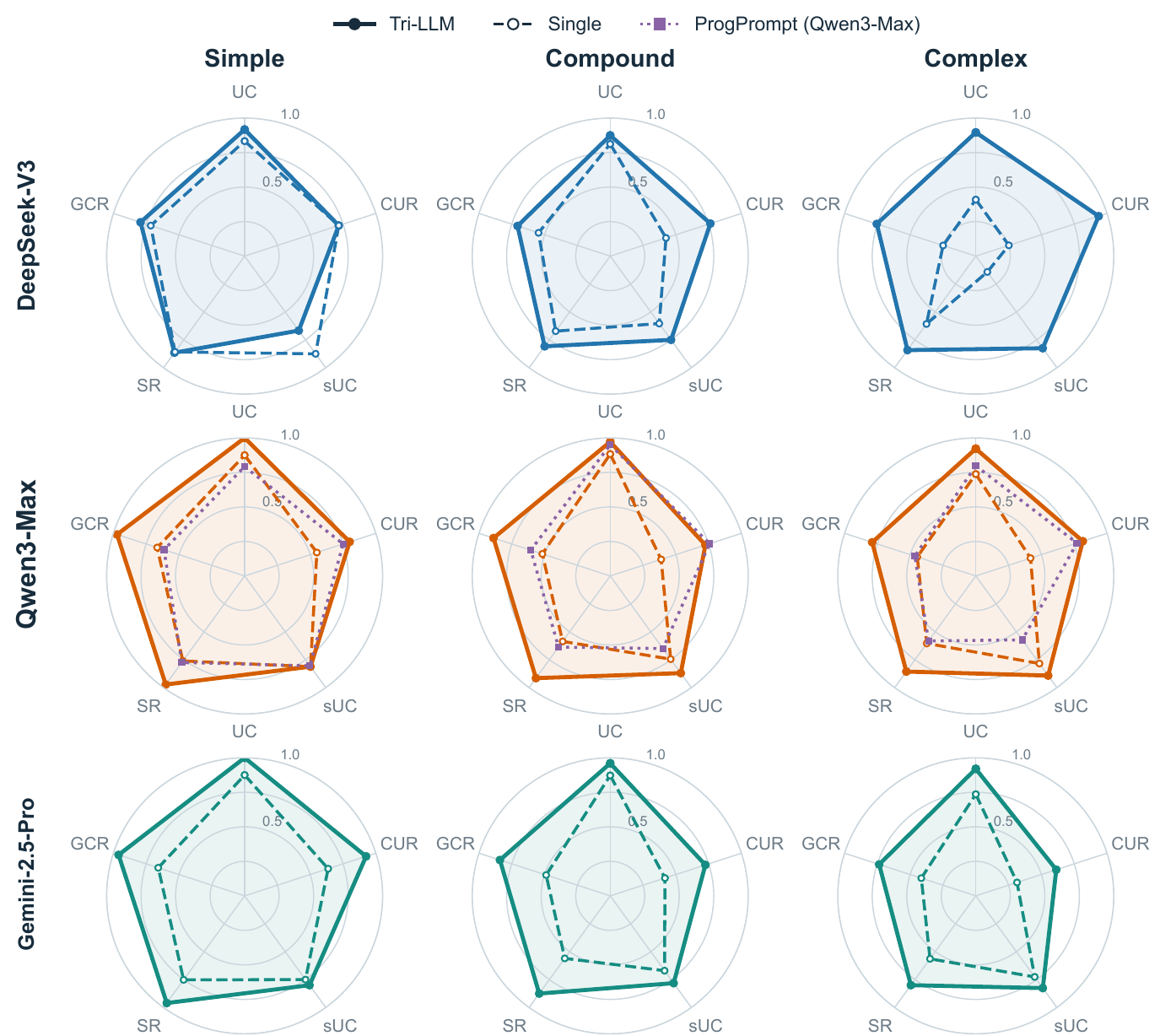}
\caption{Five-metric profiles (0--1 scale): complexity tiers by column and backbones by row. Solid/dashed profiles denote Tri/Single; dotted ProgPrompt appears only with Qwen3-Max. Exact metric values: Table~\ref{tab:main_results}.}
\label{fig:main_profiles}
\end{figure}
\begin{table*}[!t]
\centering
\caption{Task formulation and execution across three complexity tiers. Higher is better; bold values mark the best score per tier and metric.}
\label{tab:main_results}
\begingroup
\fontsize{8.2}{10}\selectfont
\setlength{\tabcolsep}{2.2pt}
\renewcommand{\arraystretch}{1.15}
\begin{tabular*}{\textwidth}{@{\extracolsep{\fill}}l*{15}{r}@{}}
\toprule
\multirow{2}{*}{Method} & \multicolumn{5}{c}{Simple} & \multicolumn{5}{c}{Compound} & \multicolumn{5}{c}{Complex} \\
\cmidrule(lr){2-6}\cmidrule(lr){7-11}\cmidrule(lr){12-16}
 & UC & CUR & sUC & SR & GCR & UC & CUR & sUC & SR & GCR & UC & CUR & sUC & SR & GCR \\
\midrule
Tri-LLM (DeepSeek-V3) & 0.917 & 0.717 & 0.667 & 0.865 & 0.793 & 0.874 & \textbf{0.763} & 0.750 & 0.808 & 0.706 & 0.896 & \textbf{0.936} & 0.825 & 0.842 & 0.754 \\
Tri-LLM (Qwen3-Max) & \textbf{1.000} & 0.800 & 0.812 & \textbf{0.972} & \textbf{0.972} & \textbf{0.972} & 0.725 & \textbf{0.870} & \textbf{0.915} & \textbf{0.889} & \textbf{0.922} & 0.815 & \textbf{0.892} & \textbf{0.856} & \textbf{0.789} \\
Tri-LLM (Gemini-2.5-Pro) & \textbf{1.000} & \textbf{0.925} & 0.800 & 0.960 & 0.960 & 0.960 & 0.725 & 0.782 & 0.875 & 0.840 & 0.920 & 0.613 & 0.826 & 0.800 & 0.736 \\
\midrule
Single-LLM (DeepSeek-V3) & 0.833 & 0.717 & \textbf{0.875} & 0.858 & 0.715 & 0.812 & 0.425 & 0.604 & 0.672 & 0.546 & 0.409 & 0.252 & 0.142 & 0.608 & 0.249 \\
Single-LLM (Qwen3-Max) & 0.875 & 0.550 & 0.812 & 0.761 & 0.666 & 0.883 & 0.388 & 0.745 & 0.586 & 0.517 & 0.738 & 0.417 & 0.784 & 0.604 & 0.446 \\
Single-LLM (Gemini-2.5-Pro) & 0.875 & 0.636 & 0.750 & 0.752 & 0.658 & 0.872 & 0.417 & 0.670 & 0.560 & 0.488 & 0.736 & 0.314 & 0.726 & 0.564 & 0.415 \\
\midrule
ProgPrompt (Qwen3-Max) & 0.792 & 0.750 & 0.804 & 0.775 & 0.614 & 0.950 & 0.757 & 0.650 & 0.637 & 0.605 & 0.798 & 0.771 & 0.572 & 0.582 & 0.464 \\
\bottomrule
\end{tabular*}
\endgroup
\end{table*}

\begin{table*}[t]
\centering
\caption{Component and policy comparisons using Qwen3-Max.}
\label{tab:controlled_results}
\begingroup
\fontsize{8.2}{9.1}\selectfont
\noindent\begin{minipage}[t]{.49\linewidth}\textbf{(a) Tri-LLM with and without Complete}\par\vspace{2pt}\fontsize{8.2}{9.1}\selectfont\setlength{\tabcolsep}{3pt}\renewcommand{\arraystretch}{1.03}\begin{tabular*}{\linewidth}{@{\extracolsep{\fill}}lrrrrrr}\toprule
&\multicolumn{3}{c}{Full}&\multicolumn{3}{c}{w/o Complete}\\ Tier&UC&CUR&sUC&UC&CUR&sUC \\ \midrule
Simple & 1.000 & .800 & .812 & .750 & .775 & .250 \\
Compound & .972 & .725 & .870 & .945 & .788 & .193 \\
Complex & .922 & .815 & .892 & .905 & .820 & .194 \\\bottomrule\end{tabular*}\end{minipage}\hfill\begin{minipage}[t]{.49\linewidth}\textbf{(b) Tri and Single with and without Audit}\par\vspace{2pt}\fontsize{8.2}{9.1}\selectfont\setlength{\tabcolsep}{3pt}\renewcommand{\arraystretch}{1.03}\begin{tabular*}{\linewidth}{@{\extracolsep{\fill}}lrrr}\toprule
Configuration&Simple&Compound&Complex \\ \midrule
Tri Full & .800 & .725 & .815 \\
Tri w/o Audit & .700 & .606 & .515 \\
Single reference & .550 & .388 & .417 \\
Single w/o Audit & .500 & .309 & .224 \\\bottomrule\end{tabular*}\end{minipage}\par\vspace{5pt}
\noindent\begin{minipage}[t]{.49\linewidth}\textbf{(c) Tri-LLM with and without functional zones}\par\vspace{2pt}\fontsize{8.2}{9.1}\selectfont\setlength{\tabcolsep}{3pt}\renewcommand{\arraystretch}{1.03}\begin{tabular*}{\linewidth}{@{\extracolsep{\fill}}lrrrrrr}\toprule
&\multicolumn{3}{c}{Full}&\multicolumn{3}{c}{w/o Zones}\\ Tier&UC&CUR&sUC&UC&CUR&sUC \\ \midrule
Simple & 1.000 & .800 & .812 & 1.000 & .530 & .725 \\
Compound & .972 & .725 & .870 & .900 & .525 & .630 \\
Complex & .922 & .815 & .892 & 1.000 & .550 & .375 \\\bottomrule\end{tabular*}\end{minipage}\hfill\begin{minipage}[t]{.49\linewidth}\textbf{(d) Continuous, failure-only, and no feedback}\par\vspace{2pt}\fontsize{8.2}{9.1}\selectfont\setlength{\tabcolsep}{3pt}\renewcommand{\arraystretch}{1.03}\begin{tabular*}{\linewidth}{@{\extracolsep{\fill}}lrrr}\toprule
Strategy&Simple&Compound&Complex \\ \midrule
Continuous & .972 & .889 & .789 \\
After failure & .970 & .802 & .783 \\
No feedback & .825 & .684 & .624 \\\bottomrule\end{tabular*}\end{minipage}\par\vspace{5pt}
\noindent\begin{minipage}[t]{.49\linewidth}\textbf{(e) One-time, selective, and exhaustive triggering}\par\vspace{2pt}\fontsize{8.2}{9.1}\selectfont\setlength{\tabcolsep}{3pt}\renewcommand{\arraystretch}{1.03}\begin{tabular*}{\linewidth}{@{\extracolsep{\fill}}lrrrrr}\toprule
Strategy&UC&sUC&SR&Task/call&Calls \\ \midrule
One-time & .732 & .731 & .768 & 10.95 & 1.0 \\
Selective HB5 & .922 & .892 & .856 & 4.45 & 3.1 \\
Selective HB10 & .903 & .783 & .820 & 4.67 & 2.9 \\
Exhaustive HB5 & .720 & .618 & .681 & 1.66 & 6.5 \\
Exhaustive HB10 & .800 & .684 & .702 & 2.50 & 4.8 \\\bottomrule\end{tabular*}\end{minipage}\hfill\begin{minipage}[t]{.49\linewidth}\textbf{(f) Tri with and without examples and zero-shot Single}\par\vspace{2pt}\fontsize{8.2}{9.1}\selectfont\setlength{\tabcolsep}{1pt}\renewcommand{\arraystretch}{1.03}\begin{tabular*}{\linewidth}{@{\extracolsep{\fill}}lrrrrrrrrr}\toprule
&\multicolumn{3}{c}{Tri examples}&\multicolumn{3}{c}{Tri zero-shot}&\multicolumn{3}{c}{Single zero-shot}\\
Tier&UC&CUR&sUC&UC&CUR&sUC&UC&CUR&sUC\\\midrule
Simple&1.000&.800&.812&.917&.725&.785&1.000&.725&.668\\
Compound&.972&.725&.870&.904&.938&.688&.723&.675&.625\\
Complex&.922&.815&.892&.666&.800&.634&.362&.558&.392\\\bottomrule
\end{tabular*}\end{minipage}\par\vspace{5pt}
\vspace{3pt}
\begin{minipage}{\textwidth}
\fontsize{8}{9.3}\selectfont\raggedright
\textit{Notes:} Full denotes Tri-LLM with all modules; w/o Audit removes Audit from the corresponding planner. Tri and Single denote Tri-LLM and Single-LLM. Trigger-policy comparisons use Complex tasks. HB denotes the heartbeat interval in seconds. With examples includes prompt demonstrations; zero-shot omits them. Higher scores are better. Calls denotes the number of complete task-formulation invocations; Task/call denotes tasks generated per invocation. Both are averaged across runs.
\end{minipage}
\endgroup
\end{table*}

Fig.~\ref{fig:main_profiles} separates goal coverage, action completeness, constraint satisfaction, and execution across backbones and complexity tiers; Table~\ref{tab:main_results} provides exact values.

On Complex tasks with Qwen3-Max, Tri-LLM reaches UC 0.922, compared with Single-LLM's 0.738 and ProgPrompt's 0.798. Its sUC of 0.892 exceeds Single-LLM by 10.8 percentage points and ProgPrompt by 32.0 points. These separate coverage gains indicate that more reference goals enter the plan and more of their required actions are specified.

Complex CUR reaches 0.815 for Qwen3-Max Tri-LLM, versus Single-LLM's 0.417. ProgPrompt reaches 0.771 despite lower step coverage and exceeds Tri-LLM on Compound CUR. Its programmatic preconditions support dependency organization, while staged completion produces more complete atomic-action sequences.

On Complex tasks with Qwen3-Max, Tri-LLM reaches SR 0.856, compared with Single-LLM's 0.604 and ProgPrompt's 0.582. Higher goal coverage and execution success yield GCR 0.789, exceeding these baselines by 34.3 and 32.5 points, respectively. The accompanying sUC gain links improved execution to more complete reference-step coverage.

With Qwen3-Max, Simple GCR is 0.972 for Tri-LLM, versus Single-LLM's 0.666 and ProgPrompt's 0.614. On Compound tasks, Tri-LLM improves GCR over Single-LLM by 37.2 points and sUC by 12.5 points, supporting intermediate-action completion across related goals.

Fig.~\ref{fig:main_profiles}'s Complex profiles show GCR gains over Single-LLM of 50.5 points with DeepSeek-V3 and 32.1 with Gemini-2.5-Pro. Tri-LLM's GCR spans 5.3 points across backbones, versus Single-LLM's 19.7. On Simple tasks, DeepSeek-V3 Single-LLM reaches sUC 0.875, compared with Tri-LLM's 0.667.

On Complex tasks, DeepSeek-V3 Tri-LLM leads CUR at 0.936, while Qwen3-Max leads sUC at 0.892. Gemini-2.5-Pro reaches UC 0.920 and CUR 0.613, separating goal coverage from dependency resolution. DeepSeek-V3 Single-LLM records sUC 0.142 and SR 0.608, distinguishing reference-step coverage from execution success.

\subsection{Completion, Dependency Checks, and Feedback}
\label{subsec:ablation}

\begin{figure*}[!t]
\centering
\includegraphics[trim=0bp 40.4bp 0bp 2.4bp,clip,width=0.90\textwidth]{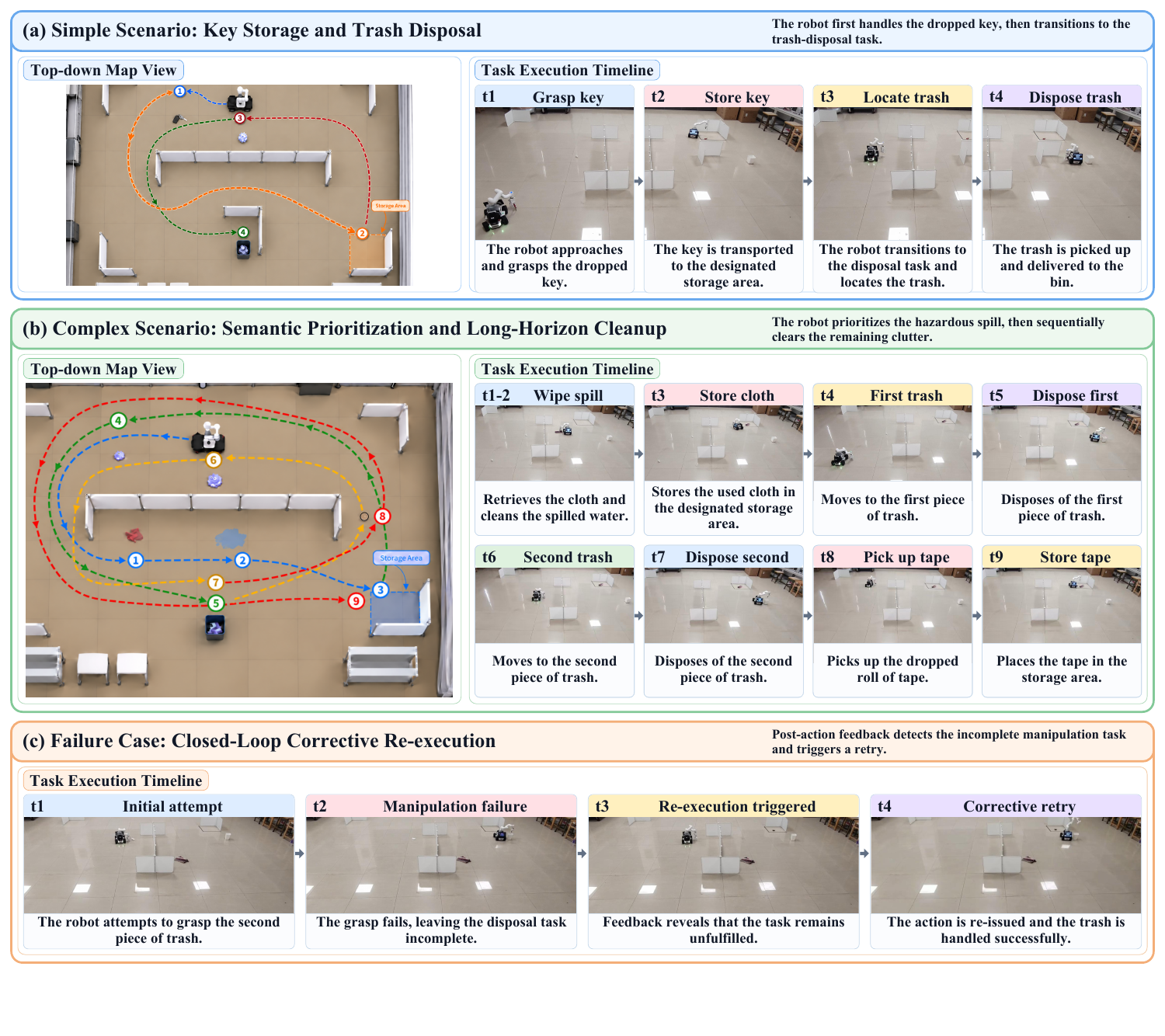}
\caption{LimoCobot validation: (a) key storage and trash disposal; (b) prioritized spill cleanup, cloth storage, two trash disposals, and tape storage; (c) failed grasp and feedback-driven retry. Routes and frames connect scene-level task ordering to selected physical movements and actions.}
\label{fig:hardware}
\end{figure*}

Complete supplies execution details for the proposed high-level goals. Removing it reduces Complex sUC from 0.892 to 0.194, a 69.8-point drop, while UC remains 0.905, close to the full model's 0.922 (Table~\ref{tab:controlled_results}(a)). Goal coverage remains high while reference-step coverage decreases sharply. The loss is also large on Simple tasks (56.2 points) and Compound tasks (67.7 points), so completion is useful even before the longest horizons. Compound CUR is 0.725 with Complete and 0.788 without it. Audit handles dependency and constraint checks as a separate reasoning role.

Audit supplies those checks. Removing it reduces Complex Tri-LLM CUR from 0.815 to 0.515; Single-LLM loses 19.3 points relative to its own reference (Table~\ref{tab:controlled_results}(b)). Both record lower CUR without Audit. Tri-LLM's loss grows from 10.0 points on Simple tasks to 11.9 on Compound tasks and 30.0 on Complex tasks, suggesting greater value at longer horizons. This contrasts with Complete's main effect on step coverage and supports retaining separate roles for filling missing actions and checking their dependencies.

Functional zones support both action grounding and constraint satisfaction. Without them, Complex sUC falls to 0.375 and CUR to 0.550, although UC rises to 1.000 (Table~\ref{tab:controlled_results}(c)). The UC, sUC, and CUR scores distinguish goal discovery from step expansion and dependency satisfaction. Spatial organization helps bind goals to destinations and supporting actions. The sUC loss grows from 8.7 points on Simple tasks to 24.0 on Compound tasks and 51.7 on Complex tasks, linking spatial grounding most strongly to detailed formulation at longer horizons.

Two simulation failure cases illustrate these mechanisms. Omitting cabinet opening before cup placement caused the cup to be deposited on the floor; leaving a cabinet door open after storage obstructed the subsequent route to a pot. The first links missing preparation steps to manipulation failure, while the second links an earlier state change to downstream navigation. Step completion, dependency auditing, and state write-back address these connected stages of execution.\par\smallskip
Feedback affects coordination after plans are generated. Compound GCR reaches 0.889 with continuous feedback, versus 0.684 without feedback and 0.802 with failure-only feedback (Table~\ref{tab:controlled_results}(d)). Continuous updates add 8.7 points beyond failure recovery. Continuous updates add 0.2 points on Simple tasks and 0.6 on Complex tasks over failure-only feedback. Relative to no feedback, the full mechanism adds 14.7 and 16.5 points in those tiers, respectively. Compound tasks show the largest additional gain from continuous rather than failure-only updates. Failure-only scores are close to continuous-feedback scores in the other tiers.

\subsection{Reasoning Admission and Prompt Examples}
\label{subsec:trigger_eval}
Selective admission records higher scores with fewer complete formulation rounds. In Complex scenarios with 15 potential tasks, selective triggering every 5~s reaches sUC 0.892, versus 0.618 for exhaustive triggering at the same interval (Table~\ref{tab:controlled_results}(e)). SR also rises from 0.681 to 0.856. Meanwhile, complete formulation invocations fall from 6.5 to 3.1, a 52.3\% reduction, and tasks per invocation rise from 1.66 to 4.45. Selective admission concentrates more tasks into each complete formulation round. Calls denotes the number of complete formulation rounds; Task/call is the mean tasks generated per round.

The interval control tests how often accumulated evidence is considered. Selective triggering every 10~s reaches sUC 0.783, above one-time formulation's 0.731, but below the 5~s policy. Moving from 10~s to 5~s adds only 0.2 formulation invocations on average while improving sUC by 10.9 points. SR also increases from 0.820 to 0.856. Recurrent admission is useful in this setting, with better plan coverage and execution at the shorter tested interval. At 10~s, selective admission still exceeds exhaustive admission in sUC (0.783 versus 0.684) while using 2.9 rather than 4.8 rounds, so the policy advantage persists across both tested intervals.

Prompt examples increase Tri-LLM step coverage, while its advantage over Single-LLM persists without examples. Without demonstrations, Complex Tri-LLM reaches sUC 0.634, versus Single-LLM's 0.392, retaining a 24.2-point gap (Table~\ref{tab:controlled_results}(f)). Adding examples raises Tri-LLM sUC to 0.892, a further gain of 25.8 points. For CUR, Complex scores change from 0.800 without examples to 0.815 with examples; Compound scores are 0.938 without examples and 0.725 with examples. Examples raise sUC across all three tiers, supporting prompt-guided atomic-step expansion. In Compound zero-shot trials, Tri-LLM also retains an 18.1-point UC advantage and a 6.3-point sUC advantage over Single-LLM. The example controls show that staged reasoning retains its formulation advantage when prompt demonstrations are removed.

\subsection{Physical Validation and Failure Recovery}
\label{subsec:hardware}
Fig.~\ref{fig:hardware}(a) shows LimoCobot storing a dropped key and disposing of trash. In panel (b), it prioritizes a water spill, cleans it with a cloth, then proceeds to cloth storage, two trash disposals, and tape storage. These runs connect scene-derived goals to physical execution without instance-specific action sequences.

Fig.~\ref{fig:hardware}(c) shows four stages of physical recovery. The robot initially attempts to grasp the second piece of trash, but the grasp fails and disposal remains incomplete. Post-action feedback identifies the unfulfilled task and updates the maintained state. The system then re-issues the action sequence; the corrective attempt grasps the trash and deposits it in the bin without a new human instruction. The observed noncompletion triggers correction before the robot advances to another goal. The sequence demonstrates feedback-driven detection of an incomplete task and corrective physical execution.

\section{Discussion and Conclusion}
CAFE links scene-driven task discovery to grounded planning and execution feedback through shared semantic memory. The experiments distinguish the roles of completion, auditing, spatial zones, and feedback. Tri-LLM records higher GCR in the tested comparisons; the controls show different sensitivities of step coverage and constraint satisfaction to module removal. Selective triggering records higher formulation and execution scores with fewer complete rounds than exhaustive triggering under the tested policies.

CAFE operates with predefined task categories and grounded robot skills, using shared semantic state to coordinate discovery and execution. Future work will enhance the generality of the system to support a wider range of scenarios and extend the framework to multi-robot coordination.

\bibliographystyle{IEEEtran}
\bibliography{references}
\end{document}